\documentclass[11pt]{article}

\usepackage[preprint]{acl}

\usepackage{times}
\usepackage{latexsym}

\usepackage[T1]{fontenc}
\usepackage[utf8]{inputenc}

\usepackage{microtype}
\usepackage{amssymb}
\usepackage{amsmath}

\usepackage{inconsolata}
\usepackage{booktabs}

\usepackage{colortbl}    % 用于 \cellcolor
\definecolor{bestcol}{RGB}{255, 200, 200}    % 浅红
\definecolor{secondcol}{RGB}{200, 220, 255}  % 浅蓝

\usepackage{graphicx}
\title{Can Segmentation Models Understand the World?\\Towards Proactive Affordance Reasoning via Visual Chain-of-Thought}

\author{
  Yuchen Guo\textsuperscript{1},
  Junli Gong\textsuperscript{2},
  Hongmin Cai\textsuperscript{3},
  Yiu-ming Cheung\textsuperscript{4},
  Weifeng Su\textsuperscript{5}
\\
  \textsuperscript{1}Northwestern University \quad
  \textsuperscript{2}Northeastern University \quad
  \textsuperscript{3}South China University of Technology
\\
  \textsuperscript{4}Hong Kong Baptist University \quad
  \textsuperscript{5}Beijing Normal - Hong Kong Baptist University
\\
  \small{
    \textbf{Correspondence:} 
    \href{mailto:yuchenguo2027@u.northwestern.edu}{yuchenguo2027@u.northwestern.edu},
    \href{mailto:wfsu@bnbu.edu.cn}{wfsu@bnbu.edu.cn}
  }
}

\begin{document}
\maketitle

\begin{abstract}

Recent segmentation models couple large language models (LLMs) with mask decoders to ground complex language expressions into masks, yet their instructions remain target-referential: they describe, constrain, or imply the region to be segmented. However, in real-world embodied interaction, human instructions are often at the intent-level, which includes the desired outcome without naming the region that enables it. To bridge this gap, we introduce \textbf{\textit{SegWorld}}, where the model reasons about the scene through a multi-level visual chain-of-thought (CoT) before committing to a mask. Before receiving any instructions, it proactively observes the scene, describing visible objects and inferring plausible events they may support.
Given an instruction, it continues the chain: from the object relevant to the intent, through the action that satisfies it, to the physical interaction site, the object part that affords the action. We formalize SegWorld as probabilistic inference, in which proactive observation supplies a linguistic scene context that improves mask prediction when instructions are given at the level of intent. We construct an intent-to-part benchmark for evaluating affordance-bearing part segmentation from high-level goals. Experiments show SegWorld matches instruction-driven baselines on target-referential instructions and improves substantially on intent-level ones.

\end{abstract}

\begin{figure*}[t!]
    \centering
    \includegraphics[width=\textwidth]{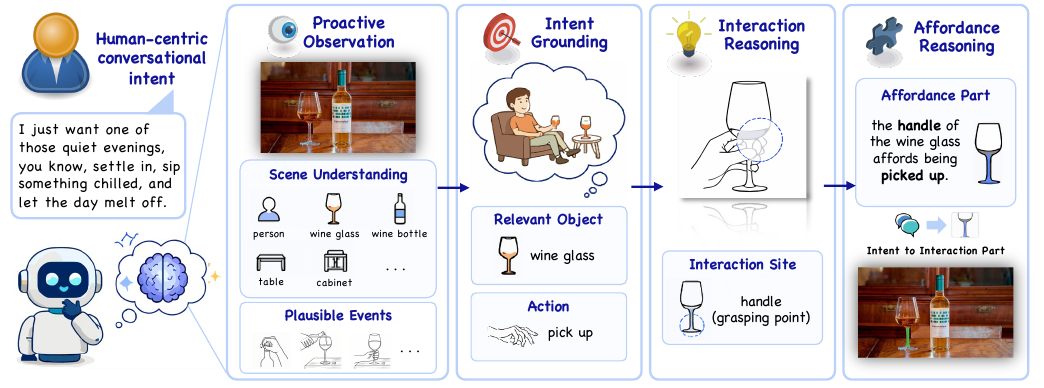}
    \caption{Given a human-centric conversational intent
that does not name the target region, SegWorld first proactively observes the
scene, then grounds the intent to a relevant object and action, reasons about the
physical interaction site, and finally segments the affordance-bearing part that
enables the intended action.}
    \label{fig:method}
    \vspace{-9pt}
\end{figure*}

\section{Introduction}
Language-driven segmentation has expanded the role of segmentation from category
recognition to instruction following. Recent models integrate LLMs or multimodal
LLMs with mask decoders, enabling them to ground complex natural-language
expressions, including spatial relations, object attributes, and implicit
reasoning cues, into masks~\citep{lai2024lisa, lu2025coprs, liu2026unipixel,
li2025sam3, gong2025reinforcing}. These instructions are commonly studied as
explicit referring expressions~\citep{Kazemzadeh, refcocog, grefcoco} or
implicit reasoning instructions~\citep{lai2024lisa, jang2025mmr,
wan2025instructpart}. Recent works further introduce visual chain-of-thought
reasoning into segmentation, using textual rationales, point or region cues, or
positional priors to connect reasoning with grounding~\citep{kao2026cot,
yuan2025sa2va}. Despite this progress, the dominant paradigm remains
query-conditioned: the model is given an instruction and is expected to return
the corresponding region. The visual scene becomes relevant only after a
question is asked, and the instruction is treated as a specification of the
target.

But this is not how human instructions work in the real world \cite{sperber1986relevance}. Human-centric instructions are pragmatic and underspecified, requiring
the listener or agent to infer the intended goal from context
\citep{shridhar2020alfred,tellex2011understanding,misra2016tell}.
Models shouldn't view scenes through the narrow aperture of current instructions. Before being asked anything, it can already take in what
is present, what could happen, and how things can be used. The instruction then
serves not as a full description of the target, but as a way of selecting one
actionable possibility from this prior understanding.

This distinction becomes critical when instructions are expressed at the level
of intent. Consider a tabletop scene with a mug, a kettle, and a plate, and the
instruction ``I want to drink water.'' The desired mask is not named by the
instruction at all. A query-conditioned model may ground it to the mug or the
kettle, both semantically related to drinking, yet neither specifies the physical
site through which the intended action should be executed. Reaching the right
region requires several steps of everyday reasoning, and the region it arrives
at is often not a whole object but a part. This is consistent with affordance and
part-level grounding studies, where interaction is localized to functional
regions rather than entire objects~\citep{do2018affordancenet, nagarajan2019grounded,
deng20213d, mo2021where2act, wan2025instructpart, tang2025uad}. We call this
setting \textit{intent-level segmentation}: recovering an affordance-bearing
part that lies several reasoning steps below a high-level goal.

We introduce \textbf{\textit{SegWorld}}, which replaces query-conditioned mapping
with a proactive, multi-level reasoning process. Rather than waiting for an
instruction, SegWorld first reads the scene on its own: it describes the visible
objects and the events they could plausibly support, producing a linguistic
scene context before any instruction arrives. When the instruction is given,
SegWorld does not search the image from scratch; it traverses this context
through an eight-level visual chain-of-thought organized into three stages. The
first stage performs proactive observation. The second identifies the object
implicated by the intent and the action that would fulfill it. The third resolves
the object part that affords this action and the affordance it carries. Each
level is expressed in language and remains explicit and inspectable, turning a
single opaque mapping from instruction to pixels into a traceable reasoning
process.

This design is principled, not merely procedural. We formalize SegWorld as probabilistic inference: the proactive stage draws a scene context, and the instruction-conditioned stages estimate the mask conditioned on both the instruction and that context. The scene context acts as an informative prior. When an instruction is target-referential, this prior matters little, since the instruction already localizes the answer. But when the instruction is intent-level, the answer is underdetermined by the instruction alone, and the scene context is what makes the posterior tractable. Proactive observation, in other words, is most valuable exactly where query-conditioned models are weakest.

% ===== Paragraph 4: Benchmark and Results =====
To study this setting, we construct \textbf{\textit{Intent2Part}}, a benchmark for
intent-level segmentation in which instructions name a high-level goal and the
target is an affordance-bearing object part. On \textit{Intent2Part}, \textit{SegWorld}
matches strong query-conditioned baselines when instructions directly specify their
target, and improves over them substantially when the target must be recovered from
intent. Beyond accuracy, its visual chain-of-thought exposes where a prediction comes
from, making both correct and incorrect masks interpretable.

\begin{figure*}[t!]
    \centering
    \includegraphics[width=\textwidth]{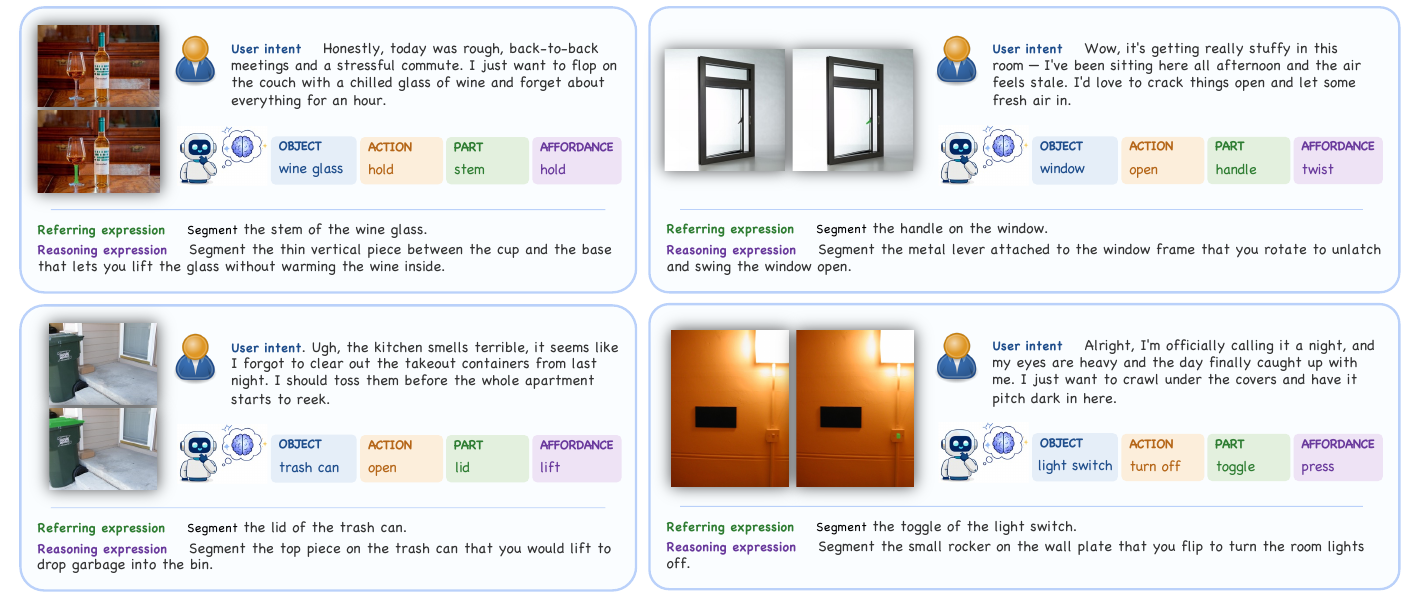}
    \caption{Examples from \textbf{\textit{Intent2Part}}. Each sample pairs a
human-centric intent-level instruction with the latent reasoning chain needed to
recover the target: object, action, part, and affordance. The benchmark also
retains target-referential referring and reasoning expressions,
allowing models to be evaluated under both direct target descriptions and
high-level intents.}
    \label{fig:method}
    \vspace{-9pt}
\end{figure*}

% ===== Paragraph 5: Contributions =====
Our contributions are summarized as follows:
\begin{itemize} 
    \item We identify \textit{\textbf{intent segmentation}}, a setting where instructions
    name a goal rather than a target, and where the answer is an affordance-bearing
    object part lying several reasoning steps below the instruction.
    \item We propose \textbf{\textit{SegWorld}}, a proactive segmentation framework that
    performs a eight-level visual chain-of-thought, and formalize it as probabilistic
    inference in which a proactive scene context serves as a prior for mask prediction.
    \item We construct \textbf{\textit{Intent2Part}}, a benchmark for evaluating
    intent-level segmentation, and show that \textit{SegWorld} matches query-conditioned
    baselines on direct instructions while improving substantially on intent-level ones.
\end{itemize}

\section{Related Work}
\label{sec:related}

\textbf{Language-driven segmentation.}
With the rapid development of LLMs, recent studies integrate LLMs into segmentation models \cite{lu2025coprs, liu2026unipixel, lai2024lisa}, enabling them to segment objects from natural-language instructions rather than single-word category labels \cite{li2025sam3, gong2025reinforcing}. Such instructions are broadly categorized into explicit referring expressions~\citep{Kazemzadeh, refcocog, grefcoco} and implicit reasoning instructions~\citep{lai2024lisa, jang2025mmr, wan2025instructpart}. To strengthen reasoning, some works introduce a visual chain-of-thought into the segmentation pipeline, using intermediate textual rationales, region or point annotations, or positional priors that couple reasoning with grounding~\citep{kao2026cot, yuan2025sa2va}. However, both referring segmentation and reasoning segmentation rely on language expressions that explicitly or implicitly describe the target region to be segmented. In contrast, our intent-level segmentation is human-centric, aiming to better serve real-world embodied agents.

\noindent\textbf{Affordance and part-level grounding.}
In real-world embodied interaction, the target of an action is often not a whole
object but the part through which the action is executed~\citep{wan2025instructpart}.
 Prior work grounds affordances in images, videos, and 3D scenes \cite{tang2025uad}, from dense affordance segmentation and graspable-region prediction to interaction hotspots and part-level affordance~\citep{do2018affordancenet, nagarajan2019grounded,
deng20213d, mo2021where2act}, with recent methods using vision-language models for open-vocabulary
affordances~\citep{qian2024affordancellm, chu20253d}. These methods take an action or interaction query as input and localize where it applies. Our setting asks the reverse: given only a high-level intent, the model infers the object, the action, and the affordance-bearing part in turn. Affordance is thus not the input query but the terminal semantics of intent-level segmentation.

\section{Intent Segmentation}
\label{sec:problem}

We address a segmentation task defined over an image $I$ and a natural-language
instruction $q$, where the goal is to predict a binary mask $M$. This setup is
shared with referring and reasoning segmentation; the distinction lies in the
nature of $q$ and the semantics of $M$.

\begin{figure*}[t!]
    \centering
    \includegraphics[width=\textwidth]{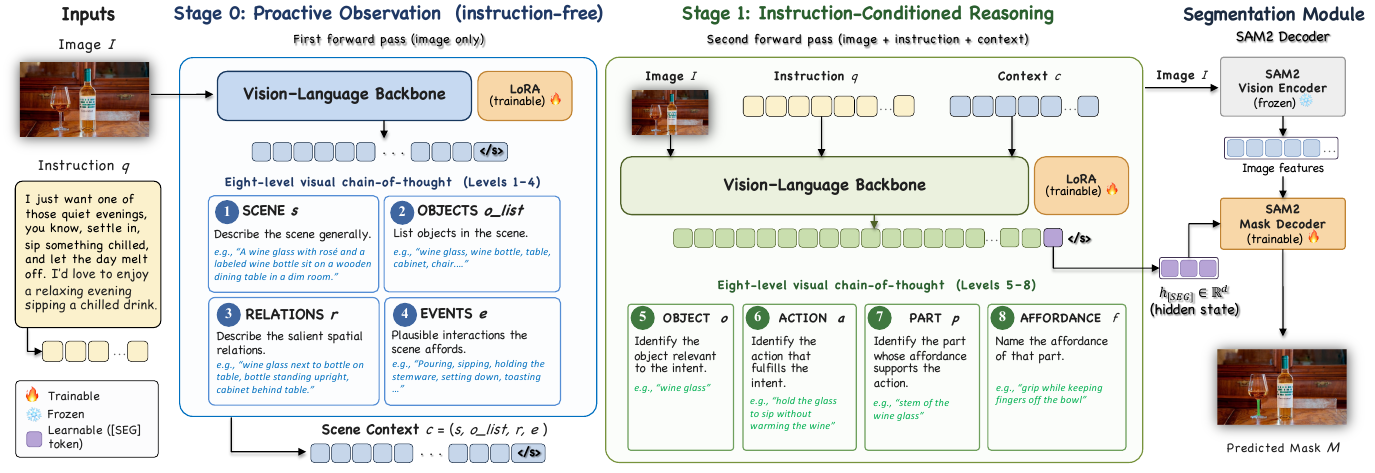}
    \caption{Overview of \textbf{\textit{SegWorld}}. Stage 0 performs instruction-free
proactive observation, producing a scene context from the image alone. Stage 1
conditions on this context and the instruction to reason through object, action,
part, and affordance. The final \texttt{[SEG]} hidden state prompts a SAM2 mask
decoder to produce the affordance part mask.}
    \label{fig:method}
    \vspace{-9pt}
\end{figure*}

In \emph{query-conditioned} segmentation, $q$ is \emph{target-referential}: it
identifies the target region by naming it, describing its attributes, or specifying
a relation that singles it out. The target is therefore determined by $I$ and $q$
together. In \emph{intent-level segmentation}, $q$ instead specifies a desired
outcome, or \emph{intent}, without referring to any image region. Because
fulfilling an intent requires acting on the scene, and physical action is
localized to a specific object part, the target $M$ is an
\emph{affordance-bearing part}: the part whose affordance enables the action that
satisfies the intent.

Satisfying $q$ in scene $I$ implicitly involves four latent quantities: an object
$o$ relevant to the intent, an action $a$ on $o$ that fulfills it, a part $p$ of
$o$ that supports the action, and the corresponding affordance $f$. The
ground-truth mask $M^{\star}$ is the segmentation of this part. Intent-level
segmentation therefore requires recovering the latent chain
$q \rightarrow o \rightarrow a \rightarrow (p,f) \rightarrow M^{\star}$, whose
intermediate terms are not given at inference time.

This structure makes intent-level segmentation harder than the query-conditioned
case. When $q$ is target-referential, the target can often be identified by
matching the instruction against image regions. When $q$ is an intent, the model
must infer which object can realize the goal, which action should be taken, and
which object part affords that action before producing a mask. This motivates the
proactive reasoning approach introduced in Section~\ref{sec:method}.

\section{The SegWorld Method}
\label{sec:method}

\subsection{Overview}

SegWorld replaces the direct mapping $q \rightarrow M$ with a two-pass,
three-stage reasoning process over a shared vision-language backbone. In the
first pass, the model observes the image $I$ without access to the instruction
and emits a linguistic scene context $c$. In the second pass, the model takes
the instruction $q$ together with this context, resolves the intent into an
object and action, grounds the action to an affordance-bearing part, and predicts
the target mask $M$. Together, the stages form an eight-level visual
chain-of-thought spanning proactive observation, intent resolution, and
part-level grounding. The chain terminates in a special \texttt{[SEG]} token
whose contextual hidden state drives a SAM2 mask decoder~\citep{ravi2025sam}.

Probabilistically, SegWorld approximates direct estimation of
$P_\theta(M \mid I,q)$ by marginalizing over an instruction-independent scene
context:
\begin{equation}
P_\theta(M \mid I,q) \approx
\mathbb{E}_{c \sim P_\theta(c \mid I)}
\big[P_\theta(M \mid I,q,c)\big].
\label{eq:marginal}
\end{equation}
The first pass models $P_\theta(c \mid I)$ and the second models
$P_\theta(M \mid I,q,c)$. In practice, we instantiate the expectation with one
decoded context from the observation pass and predict the mask conditioned on
that context.

\subsection{Proactive Observation}
\label{subsec:observation}

The first stage models the instruction-free distribution $P_\theta(c \mid I)$.
The instruction $q$ is withheld entirely: the observation pass receives only the
image and an observation prompt, ensuring that the resulting context is a
property of the scene rather than of the query.

We decompose the context into four levels,
$c=(s,o_{\text{list}},r,e)$. The \textsc{Scene} level gives a one-sentence
description $s$ of the overall setting. The \textsc{Objects} level enumerates
salient visible objects $o_{\text{list}}$. The \textsc{Relations} level captures
spatial and functional relations $r$ among them. The \textsc{Events} level
predicts a small set of plausible interactions $e$ supported by the scene,
independent of any particular instruction. These levels summarize what is
present, how it is arranged, and what actions the scene could plausibly afford
before any goal is specified.

\subsection{Instruction-Conditioned Reasoning}
\label{subsec:reasoning}

Given the image $I$, instruction $q$, and scene context $c$, the second pass
models $P_\theta(M \mid I,q,c)$ through two further stages. The intent-resolution
stage identifies the object $o$ relevant to the intent and the action $a$ that
would satisfy it. The affordance-grounding stage identifies the part $p$ of $o$
that supports the action and names its affordance $f$. These four levels recover
the latent chain introduced in Section~\ref{sec:problem},
$q \rightarrow o \rightarrow a \rightarrow (p,f) \rightarrow M^\star$, making the
intermediate reasoning explicit rather than collapsing the instruction directly
into a mask.

The final token of the chain is \texttt{[SEG]}, added to the backbone vocabulary.
Its contextual hidden state $h_{\text{[SEG]}}\in\mathbb{R}^{d}$ is linearly
projected into a prompt embedding for the SAM2 mask decoder. The decoder fuses
this prompt with image features from a frozen SAM2 vision encoder to produce the
binary mask $M$. Thus, the mask decoder receives a learned language prompt that
encodes the full reasoning chain, rather than the raw instruction alone.

\paragraph{Why the scene context helps.}
Equation~\ref{eq:marginal} clarifies when proactive observation is useful. For
target-referential instructions, $q$ already identifies the target, so additional
scene context has limited effect. For intent-level instructions, however,
$P_\theta(M \mid I,q)$ is diffuse: multiple objects and parts may be semantically
related to the intent, but only some are actionable in the observed scene.
Conditioning on $c$ restricts prediction to candidates consistent with the scene
structure and plausible events, sharpening the posterior. We treat
Equation~\ref{eq:marginal} as a modeling approximation and validate its utility
empirically in Section~\ref{sec:experiments}.

\subsection{Training}
\label{subsec:training}

\paragraph{Objective.}
SegWorld is jointly trained with mask and language-modeling objectives:
\begin{equation}
\begin{aligned}
\mathcal{L} =\;&
\lambda_{\text{mask}}\mathcal{L}_{\text{mask}}(M,M^\star)
+ \lambda_0 \mathcal{L}_{\text{LM}}^{(0)}(c) \\
&+ \lambda_1 \mathcal{L}_{\text{LM}}^{(1)}(o,a,p,f),
\end{aligned}
\label{eq:loss}
\end{equation}
where $\mathcal{L}_{\text{mask}}$ is binary cross-entropy plus dice loss, and
$\mathcal{L}_{\text{LM}}^{(0)}$ and $\mathcal{L}_{\text{LM}}^{(1)}$ supervise the
observation chain $c=(s,o_{\text{list}},r,e)$ and the instruction-conditioned
chain $(o,a,p,f)$, respectively. We use
$\lambda_{\text{mask}}=1.0$, $\lambda_0=0.5$, and $\lambda_1=1.0$. The mask loss
back-propagates through the \texttt{[SEG]} hidden state into the second pass,
while the observation pass is supervised by its language-modeling loss.

\paragraph{Supervision.}
The instruction-conditioned levels $(o,a,p,f)$ are taken from the ground-truth
annotations of \textit{Intent2Part} (Section~\ref{sec:benchmark}). The observation
levels $(s,o_{\text{list}},r,e)$ are synthesized offline with a vision-language
model using an image-only prompt, so they describe the scene without reference to
any instruction. During training, the instruction $q$ is sampled from a pool
containing the target-referential instructions inherited from InstructPart and
our intent-level instruction, exposing the model to both direct and intent-level
query distributions.

\paragraph{Scheduled sampling.}
At inference, the second pass conditions on the context generated by the first
pass. During early training, this context is noisy, so we first condition on the
synthesized observation context and gradually replace it with self-generated
contexts after a warmup of $S_{\text{warmup}}$ steps. The replacement probability
increases linearly up to $p_{\max}=0.5$. Self-generated contexts are detached
before entering the second pass, while the observation pass is always supervised
by its language-modeling loss. This curriculum reduces the train-test gap while
preserving early optimization stability.

\paragraph{Parameterization.}
The vision-language backbone is fine-tuned with LoRA adapters applied to
attention and MLP projections. The SAM2 vision encoder is frozen, and the SAM2
mask decoder is trained. Full implementation details and hyperparameters are
reported in Appendix~\ref{appendix:training}.

\section{The Intent2Part Benchmark}
\label{sec:benchmark}

To evaluate intent-level segmentation, we construct
\textbf{\textit{Intent2Part}}, a benchmark in which instructions express
high-level intents and the target is an affordance-bearing object part.
Intent2Part builds on InstructPart~\citep{wan2025instructpart}, which provides images,
object and part labels, part affordances, high-level actions, part masks, and
target-referential instructions. We retain these annotations and add one
intent-level instruction for each image-annotation pair.

\subsection{Construction}
\label{subsec:construction}

\begin{figure*}[t!]
    \centering
    \includegraphics[width=\textwidth]{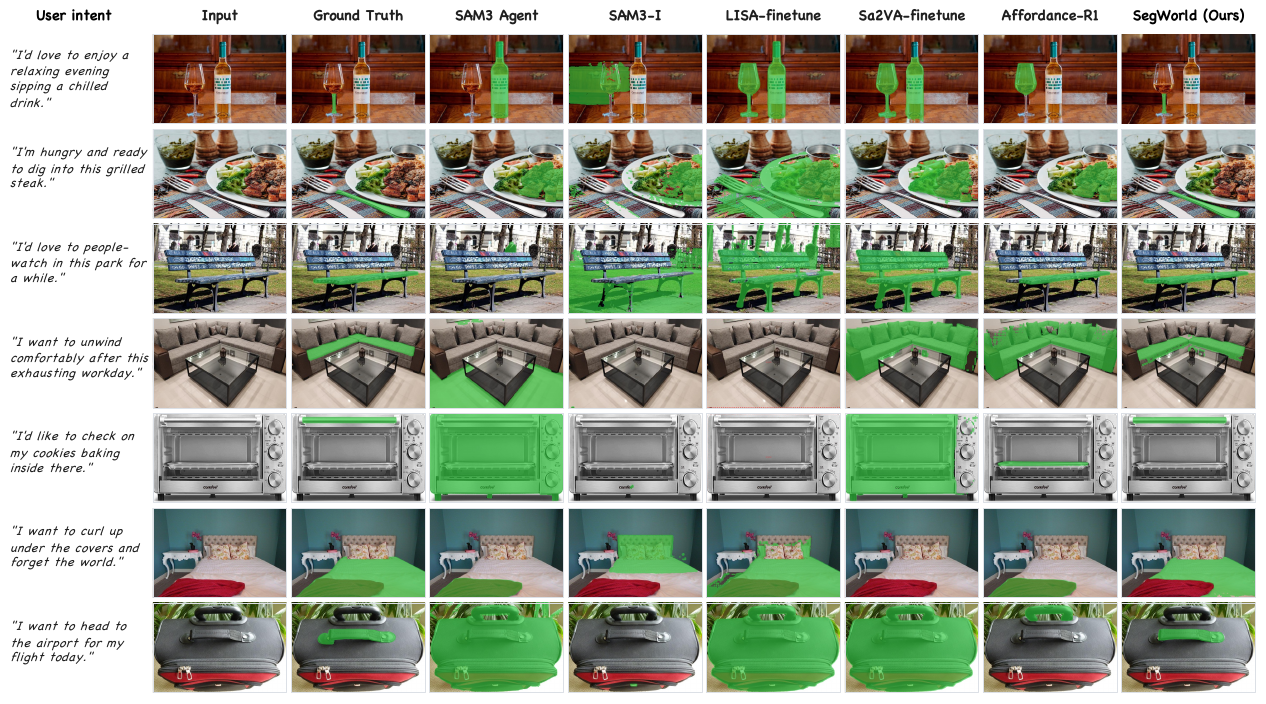}
    \caption{Qualitative comparison on intent-level instructions. Existing
    methods often ground the intent to a semantically related whole object or
    broad region, while SegWorld localizes the affordance-bearing part needed to
    realize the intent, such as a bottle body for pouring, a bench seat for
    sitting, an oven handle for opening, or a suitcase handle for carrying.}
    \label{fig:qualitative}
    \vspace{-9pt}
\end{figure*}

\paragraph{Intent-level instructions.}
For each sample, we author an instruction that names only the desired outcome,
without mentioning the target object, target part, action, or affordance. To
reduce language and stylistic bias from any single generator, we generate
candidate instructions independently with three frontier vision-language models:
\texttt{Claude Opus 4.7}, \texttt{GPT-5.5}, and \texttt{Gemini 3.1 Pro}. Each
model is prompted in a first-person voice with the image and a brief description
of a plausible user scenario. We then filter candidates with an automatic
validator that enforces first-person intent form, length constraints, and
exclusion of target-revealing terms and their variants. Candidates that fail
validation are rewritten manually. The full generation prompt, validator, and
quality-control procedure are given in Appendix~\ref{appendix:benchmark}.

\paragraph{Observation-level annotations.}
To supervise SegWorld's proactive observation stage, we additionally synthesize
image-level scene contexts for training images. Given an image without any
instruction, a vision-language model produces a scene description, salient
objects, spatial and functional relations, and plausible events. These
annotations supervise the observation pass during training and are never used for
evaluation.

\subsection{Statistics and Splits}
\label{subsec:statistics}

Intent2Part contains 1{,}800 training and 600 test image-annotation pairs,
following the splits of InstructPart, and adds intent-level instructions to both
training and test samples. We evaluate on two test splits:
\textbf{test\_official}, the original 600-sample InstructPart test split, and
\textbf{test\_clean}, a 232-sample subset whose base images do not appear in
training. Since \textbf{test\_official} contains repeated base images with
different part annotations, \textbf{test\_clean} serves as our primary split for
measuring image-level generalization. Each sample is evaluated under both the
inherited target-referential instruction and the newly authored intent-level
instruction.

\section{Experiments}
\label{sec:experiments}

\subsection{Setup}
\label{subsec:setup}

\paragraph{Datasets and metrics.}
All experiments are conducted on \textit{Intent2Part}
(Section~\ref{sec:benchmark}). Models are trained on the 1{,}800-sample training
split and evaluated under two instruction conditions: \emph{target-referential}
instructions inherited from InstructPart and our newly authored
\emph{intent-level} instructions. Intent-level evaluation is reported on
\textit{test\_clean} (232 source-disjoint samples; primary split) and
\textit{test\_official} (600 samples). We report mean Intersection-over-Union
(mIoU), cumulative IoU (cIoU), and the \texttt{[SEG]} emission rate, defined as
the fraction of samples for which the model produces a valid segmentation token.

\begin{table*}[t]
\centering
\small
\setlength{\tabcolsep}{4pt}
\renewcommand{\arraystretch}{1.1}
\begin{tabular}{lcccccccc|ccc|c}
\toprule
& \multicolumn{2}{c}{\textbf{Referring}} 
& \multicolumn{2}{c}{\textbf{Reasoning}} 
& \multicolumn{4}{c}{\textbf{Intent-level}} 
& \multicolumn{3}{c}{\textbf{Per-Action mIoU}} 
& \\
\cmidrule(lr){2-3} \cmidrule(lr){4-5} \cmidrule(lr){6-9} \cmidrule(lr){10-12}
& \multicolumn{2}{c}{\textit{InstructPart}}
& \multicolumn{2}{c}{\textit{InstructPart}}
& \multicolumn{2}{c}{\textit{test\_clean}} 
& \multicolumn{2}{c}{\textit{test\_official}}
& \multicolumn{3}{c}{\textit{Intent, test\_official}} 
& \\
\cmidrule(lr){2-3} \cmidrule(lr){4-5} \cmidrule(lr){6-7} \cmidrule(lr){8-9} \cmidrule(lr){10-12}
\textbf{Method} 
& mIoU & cIoU 
& mIoU & cIoU 
& mIoU & cIoU 
& mIoU & cIoU 
& hold & cut & sit 
& \textbf{[SEG]\%} \\
\midrule
vanilla Sa2VA      
  & 0.510 & 0.367 
  & 0.498 & 0.359 
  & 0.151 & 0.099 
  & 0.141 & 0.093 
  & 0.178 & 0.073 & 0.075 
  & 57.0 \\

SAM3 Agent         
  & 0.592 & 0.426 
  & 0.580 & 0.418 
  & 0.387 & 0.252 
  & 0.366 & 0.238 
  & 0.150 & 0.502 & 0.548 
  & 98.0 \\

LISA-finetune      
  & 0.583 & 0.420 
  & 0.572 & 0.412 
  & 0.221 & 0.144 
  & 0.205 & 0.133 
  & 0.155 & 0.205 & 0.265 
  & \textbf{100.0} \\

UniPixel-finetune      
  & 0.627 & 0.451 
  & 0.615 & 0.443 
  & 0.253 & 0.164 
  & 0.241 & 0.157 
  & 0.162 & 0.290 & 0.335 
  & \textbf{100.0} \\

Affordance-R1
  & 0.685 & 0.493 
  & 0.672 & 0.484 
  & 0.480 & 0.313 
  & 0.420 & 0.273 
  & 0.175 & 0.610 & 0.660 
  & 99.0 \\

SAM3-I             
  & 0.701 & 0.505 
  & 0.689 & 0.496 
  & 0.342 & 0.222 
  & 0.318 & 0.207 
  & 0.165 & 0.445 & 0.498 
  & \textbf{100.0} \\

Sa2VA-finetune     
  & 0.712 & 0.513 
  & 0.703 & 0.506 
  & 0.305 & 0.198 
  & 0.283 & 0.184 
  & 0.170 & 0.390 & 0.410 
  & 92.0 \\
\midrule

SegWorld v1        
  & 0.782 & 0.563 
  & 0.770 & 0.554 
  & 0.514 & 0.345 
  & 0.446 & 0.310 
  & 0.184 & 0.655 & 0.729 
  & \textbf{100.0} \\

\textbf{SegWorld v2} 
  & \textbf{0.783} & \textbf{0.564} 
  & \textbf{0.772} & \textbf{0.556} 
  & \textbf{0.612} & \textbf{0.411} 
  & \textbf{0.495} & \textbf{0.345} 
  & \textbf{0.253} & \textbf{0.734} & \textbf{0.775} 
  & \textbf{100.0} \\
\bottomrule
\end{tabular}
\caption{Main results on \textit{Intent2Part}. We report mIoU/cIoU across
target-referential and intent-level instructions, with \textit{test\_clean} as
the primary intent-level split. Per-action mIoU covers three representative
actions, and \textbf{[SEG]\%} denotes valid mask-token emission. Best results are
in \textbf{bold}.}
\vspace{-9pt}
\label{tab:main}
\end{table*}

\paragraph{Baselines.}
We compare with representative reasoning and affordance segmentation baselines:
LISA~\citep{lai2024lisa}, Sa2VA~\citep{yuan2025sa2va},
Affordance-R1~\citep{wang2026affordance}, and SAM3-I~\citep{li2025sam3}. We also
include a frozen \textbf{SAM3 Agent} cascade that uses Qwen-VL~\citep{bai2025qwen3} to
translate the instruction into a target description and SAM3~\citep{carion2025sam} to
predict the mask. Trainable baselines are fine-tuned on \textit{Intent2Part}
with the same data and training budget as SegWorld. We additionally report
off-the-shelf Sa2VA without task-specific fine-tuning.

\paragraph{SegWorld variants.}
\textbf{SegWorld v1} is trained only with target-referential instructions and is
evaluated zero-shot on intent-level instructions. \textbf{SegWorld v2} adds our
intent-level instructions to the training pool with 20\% sampling probability.
Implementation details are provided in Appendix~\ref{appendix:training}.

\subsection{Qualitative Examples}
\label{subsec:qualitative}

Figure~\ref{fig:qualitative} compares SegWorld with representative baselines on
intent-level instructions. Existing methods often identify objects that are
semantically related to the intent but fail to isolate the actionable part:
bottles instead of pourable regions, whole benches or couches instead of seats,
and entire appliances or luggage instead of handles. SegWorld more consistently
grounds the intent to the affordance-bearing part, reflecting the benefit of
explicitly reasoning through object, action, part, and affordance before emitting
the segmentation token.

\subsection{Main Results}
\label{subsec:main_results}

Table~\ref{tab:main} quantifies the trends shown in
Figure~\ref{fig:qualitative}, comparing SegWorld with existing segmentation
systems across target-referential and intent-level instructions.

\paragraph{SegWorld preserves target-referential segmentation.}
On the standard referring and reasoning conditions, SegWorld v1 already improves
over the strongest fine-tuned baseline, and v2 maintains this performance after
adding intent-level supervision. This shows that proactive observation and
multi-level reasoning do not trade off against conventional instruction-driven
segmentation.

\paragraph{Proactive reasoning transfers to intent-level instructions.}
Without seeing intent-level instructions during training, SegWorld v1 achieves
0.514 mIoU on intent-level \textit{test\_clean}, outperforming the strongest baseline, Affordance-R1, by 3.4 points. Since v1 is trained only on
target-referential instructions, this gain indicates that the two-pass reasoning
structure learns a reusable intent-to-part inference pattern rather than merely
memorizing intent templates.

\paragraph{Intent supervision further improves intent-to-part grounding.}
Adding intent-level instructions in SegWorld v2 raises mIoU on
\textit{test\_clean} from 0.514 to 0.612, while leaving target-referential
performance essentially unchanged. Compared with Affordance-R1, which applies
MLLM-based affordance reasoning to this setting, SegWorld v2 improves
intent-level mIoU on \textit{test\_clean} from 0.253 to 0.612. Although
\textit{test\_clean} is source-disjoint from training, it yields higher scores
than \textit{test\_official}; Section~\ref{subsec:analysis} shows that the
overlapping-image subset of \textit{test\_official} contains conflicting part
annotations and is empirically harder.

\begin{figure*}[t!]
    \centering
    \includegraphics[width=\textwidth]{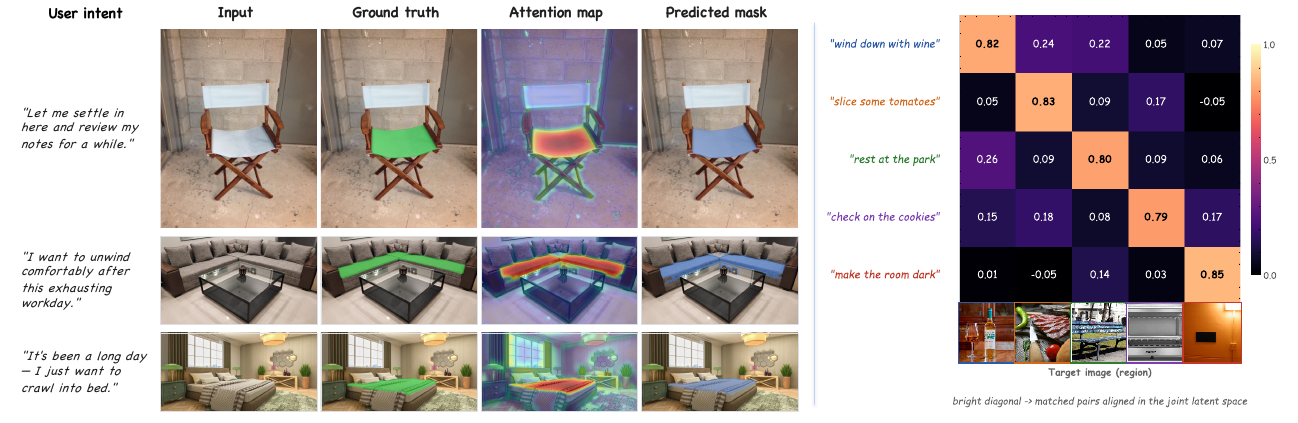}
    \caption{Reasoning visualization for SegWorld. Left: attention maps focus
    on the actionable part before mask prediction, and the predicted masks align
    with the affordance-bearing regions required by the intent. Right:
    intent-region similarity scores form a bright diagonal, showing that matched
    intent-region pairs align in the learned joint space.}
    \label{fig:heat}
    \vspace{-9pt}
\end{figure*}

\subsection{Analysis}
\label{subsec:analysis}

\paragraph{Reasoning aligns with actionable parts.}
Figure~\ref{fig:heat} visualizes SegWorld's grounding behavior. The grounding
heatmaps concentrate on the part that realizes the intent before the final mask
is produced, such as a chair seat, sofa cushion, or bed surface. The
intent-region similarity matrix shows a bright diagonal, indicating that matched
intent-image pairs align in the learned joint space. These visualizations suggest
that SegWorld's gains come from reasoning toward affordance-bearing parts rather
than merely selecting semantically related objects.

\paragraph{Two-pass training induces mask commitment.}
On intent-level \textit{test\_official}, off-the-shelf Sa2VA fails to emit
\texttt{[SEG]} on 43.0\% of samples, often responding conversationally instead
of producing a mask. Both SegWorld variants emit valid \texttt{[SEG]} tokens on
all samples. This suggests that the two-pass training procedure changes the
model's behavior from answering an intent in language to committing the intent
to a segmentation output.

\paragraph{Image overlap introduces part conflict.}
Although \textit{test\_clean} is source-disjoint from training, it yields higher
intent-level scores than \textit{test\_official}. The latter includes 368
overlapping-image samples whose base images appear in training with different
object-part annotations. These samples are harder: their implied mIoU is 0.403
for SegWorld v1 and 0.422 for v2, both below the corresponding
\textit{test\_clean} scores. The gap therefore reflects part-level conflict, not
image memorization.

\begin{table}[t]
\centering
\small
\setlength{\tabcolsep}{4pt}
\renewcommand{\arraystretch}{1.05}
\begin{tabular}{lcccc}
\toprule
\textbf{Variant} 
& \multicolumn{2}{c}{\textit{test\_clean}} 
& \multicolumn{2}{c}{\textit{test\_official}} \\
\cmidrule(lr){2-3} \cmidrule(lr){4-5}
& mIoU & cIoU & mIoU & cIoU \\
\midrule
SegWorld v2 
& \textbf{0.612} & \textbf{0.411} 
& \textbf{0.495} & \textbf{0.345} \\
w/o intent supervision (v1)
& 0.514 & 0.345
& 0.446 & 0.310 \\
w/o event level
& 0.573 & 0.386
& 0.472 & 0.329 \\
w/o proactive context
& 0.438 & 0.294
& 0.382 & 0.266 \\
w/o Stage-1 CoT
& 0.402 & 0.270
& 0.358 & 0.249 \\
\bottomrule
\end{tabular}
\caption{
Ablation on intent-level segmentation. We report mIoU/cIoU on both intent-level
splits. Removing either proactive observation or instruction-conditioned
reasoning degrades intent-to-part grounding.
}
\label{tab:ablation}
\vspace{-9pt}
\end{table}

\paragraph{Per-action behavior.}
Per-action results, reported in Appendix~\ref{appendix:per_action}, show that
intent supervision is especially helpful for small interaction parts such as
handles, knobs, and straps, where the intended object is easy to identify but the
actionable part is spatially small.

\paragraph{Ablation.}
Table~\ref{tab:ablation} isolates the contribution of SegWorld's reasoning
components. Removing the event level weakens performance, indicating that
plausible scene interactions provide useful priors beyond static scene
description. Removing the proactive context causes a larger drop, showing that
instruction-free observation is a key source of intent-level grounding. The
largest degradation comes from removing the Stage-1 chain, confirming that
explicit object-action-part-affordance reasoning is not merely explanatory text
but contributes directly to mask prediction.

\section{Conclusion}

We introduced \textbf{\textit{SegWorld}}, a proactive framework for intent-level
segmentation. SegWorld first observes the scene without an instruction, then
reasons from a high-level intent to an object, action, affordance-bearing part,
and mask through a visual chain-of-thought. We also introduced
\textbf{\textit{Intent2Part}}, a benchmark for evaluating part segmentation from
high-level goals. Experiments show that SegWorld maintains strong
target-referential performance while substantially improving intent-level
grounding, suggesting a step toward segmentation systems that reason about what
a scene affords rather than only following instructions.

\section*{Limitations}

SegWorld studies intent-level segmentation in a controlled part-segmentation
setting. First, Intent2Part focuses on static images and single-step intents.
Many embodied scenarios involve temporal context, multi-step plans, or changing
object states, which are outside the scope of this work. Second, our
intent-level instructions are authored with the aid of frontier vision-language
models and filtered by automatic validators. Although this reduces direct target
leakage and single-model language bias, the resulting language distribution may
still differ from naturally collected human instructions. Third, SegWorld makes
its reasoning explicit through language, which improves inspectability but also
makes the system sensitive to the quality of intermediate textual predictions:
errors in scene context, object selection, or action inference can propagate to
the final mask. Finally, the benchmark defines one ground-truth
affordance-bearing part per intent, while real scenes may admit multiple valid
ways to satisfy the same goal. Our evaluation therefore does not measure multi-solution intent grounding.

\section*{Ethical Considerations}

SegWorld is designed for intent-level part segmentation and may be useful for
assistive interfaces, embodied agents, and human-centered visual grounding. As
with other vision-language systems, its predictions can be affected by biases in
the underlying models and datasets, including cultural assumptions about common
goals, household objects, and typical object use. Incorrect intent grounding may
also lead to inappropriate or unsafe action suggestions if used directly in a
robotic system. Our work only predicts segmentation masks and does not execute
actions; any deployment in embodied settings should include human oversight,
task-specific safety checks, and evaluation under the target environment. The
intent-level instructions in Intent2Part are generated and filtered to avoid
revealing target regions, but they may not fully reflect the diversity of
naturally occurring human language. We will release the dataset construction
protocol and validation rules to support reproducibility and auditing.

\bibliography{custom}

\appendix

\section{Training and Implementation Details}
\label{appendix:training}

\paragraph{Model.}
SegWorld is built on Sa2VA-Qwen2.5-VL-7B. The vision-language backbone is
fine-tuned with LoRA adapters applied to attention and MLP projections, with
rank $r=64$, $\alpha=128$, and dropout $0.05$. The SAM2 vision encoder is frozen
throughout training, while the SAM2 mask decoder and the projection from the
\texttt{[SEG]} hidden state to the decoder prompt embedding are trained.

\paragraph{Optimization.}
We train with AdamW using learning rate $2\times10^{-5}$, weight decay $0.05$,
bf16 precision, and DeepSpeed ZeRO-2. The effective batch size is 16, with batch
size 1 per device, 8-step gradient accumulation, and 2 A100-80G GPUs. Each run is
trained for 2{,}250 optimizer steps and takes approximately 4 hours. Gradient
checkpointing is disabled to avoid double gradient reduction across the two-pass
forward computation.

\paragraph{Scheduled sampling.}
During inference, Stage 1 conditions on the scene context generated by Stage 0.
During early training, Stage 1 is conditioned on the synthesized Stage-0 context.
After a warmup of $S_{\text{warmup}}$ steps, we gradually replace the synthesized
context with the model's self-generated context. The replacement probability is
$p_{\text{self}}(t)=\min(t/S_{\text{warmup}},1)\cdot p_{\max}$, with
$p_{\max}=0.5$. Self-generated contexts are decoded under \texttt{no\_grad} and
detached before entering Stage 1. Stage 0 is always supervised with its
language-modeling loss.

\paragraph{Baseline training.}
Trainable baselines are fine-tuned on the same Intent2Part training split with
the same instruction pool, optimizer, batch size, training budget, and mask loss
as SegWorld whenever their released implementations support supervised
fine-tuning. Off-the-shelf Sa2VA is evaluated without task-specific fine-tuning.
SAM3 Agent is a frozen cascade: Qwen-VL rewrites the instruction into a target
description, and SAM3 predicts the mask from that description.

\section{Intent2Part Construction Details}
\label{appendix:benchmark}

\paragraph{Intent-level instruction generation.}
For each image-annotation pair, we generate candidate intent-level instructions
with three frontier vision-language models: \texttt{Claude Opus 4.7},
\texttt{GPT-5.5}, and \texttt{Gemini 3.1 Pro}. Each model receives the image and
a brief description of a plausible user scenario, and is prompted to write a
first-person instruction that expresses a desired outcome without naming the
target object, part, action, or affordance. Using multiple generators reduces
language and stylistic bias from any single model.

\paragraph{Validation.}
Each candidate instruction is checked by an automatic validator. The validator
requires a first-person intent form, length between 6 and 25 words, and absence
of the ground-truth \texttt{object}, \texttt{part}, \texttt{action}, and
\texttt{affordance} terms. It also filters plural forms, common morphological
variants, and near-synonyms that would reveal the target region. If multiple
candidates pass validation, we select the most natural one. If no candidate
passes, the instruction is manually rewritten and re-validated. Every released
intent-level instruction passes the validator.

\paragraph{Observation-level annotations.}
Stage-0 supervision is synthesized with a vision-language model using an
image-only prompt. The model is asked to produce four fields: a one-sentence
scene description, a list of salient visible objects, spatial and functional
relations among them, and plausible events supported by the scene. The prompt
does not include any target-referential or intent-level instruction. These
annotations are used only for training the proactive observation stage and are
never used during evaluation.

\paragraph{Split construction.}
Intent2Part follows the original InstructPart train/test split, with 1{,}800
training and 600 test image-annotation pairs. Because InstructPart may contain
multiple annotations for the same base image, the official test split includes
some images whose base image also appears in training with a different
object-part annotation. We therefore define \textit{test\_clean} as the subset of
test samples whose base image does not appear in training. This yields 232
source-disjoint test samples and 368 overlapping-image test samples. We report
both \textit{test\_official} and \textit{test\_clean}, using
\textit{test\_clean} as the primary split for image-level generalization.

\section{Per-Action Results}
\label{appendix:per_action}

Table~\ref{tab:per_action_app} reports intent-level mIoU on
\textit{test\_official} for the eight most frequent action categories. SegWorld
v1 already improves substantially over off-the-shelf Sa2VA on actions requiring
localized part grounding, while v2 further improves categories where the
actionable part is small relative to the object, such as \textit{pick up} and
\textit{open}.

\begin{table}[h]
\centering
\small
\setlength{\tabcolsep}{5pt}
\begin{tabular}{lrrrr}
\toprule
\textbf{Action} & $n$ & vanilla & v1 & v2 \\
\midrule
hold     & 93 & 0.178 & 0.184 & \textbf{0.253} \\
cut      & 72 & 0.073 & 0.655 & \textbf{0.734} \\
sit      & 71 & 0.075 & 0.729 & \textbf{0.775} \\
pick up  & 66 & 0.101 & 0.152 & \textbf{0.482} \\
open     & 50 & 0.020 & 0.455 & \textbf{0.565} \\
support  & 44 & 0.234 & \textbf{0.459} & 0.413 \\
grasp    & 33 & 0.147 & 0.379 & \textbf{0.405} \\
contain  & 32 & 0.405 & 0.402 & \textbf{0.627} \\
\bottomrule
\end{tabular}
\caption{Per-action mIoU on intent-level \textit{test\_official}.}
\label{tab:per_action_app}
\end{table}

\end{document}